\documentclass[10pt,twocolumn,letterpaper]{article}

\usepackage{cvpr}              

\DeclareMathOperator*{\argmax}{arg\,max}

\definecolor{cvprblue}{rgb}{0.21,0.49,0.74}
\usepackage[pagebackref,breaklinks,colorlinks,allcolors=cvprblue]{hyperref}
\usepackage{booktabs}
\usepackage{multirow}
\usepackage[table]{xcolor} 
\usepackage{adjustbox} 
\usepackage[T1]{fontenc}
\AddToHook{shipout/foreground}{%
  \put(0,-26){%
    \makebox[\paperwidth]{\sffamily\bfseries\footnotesize\color{gray} PREPRINT}%
  }%
}
\def\paperID{*****} 
\def\confName{CVPR}
\def\confYear{2026}

\title{
RhythmJEPA: Rhythm-Structured Predictive Learning
for \\ Remote Photoplethysmography
}

\author{
{\large
Ba-Thinh Nguyen,
Huu-Dung Nguyen,
Thi-Duyen Ngo,
Thanh-Ha Le$^{*}$,
}\\[0.6em]
VNU University of Engineering and Technology, Hanoi, Vietnam\\
[0.4em]
{\tt\small
\{22028163, 22028076, duyennt, ltha\}@vnu.edu.vn
}
}

\begin{document}
\maketitle

\begingroup
\renewcommand{\thefootnote}{\fnsymbol{footnote}}
\setcounter{footnote}{0}
\footnotetext[1]{Corresponding author: \href{mailto:ltha@vnu.edu.vn}{ltha@vnu.edu.vn}.}
\endgroup
\begin{abstract}
Remote photoplethysmography (rPPG) estimates physiological signals from facial videos by analyzing subtle pulse-induced skin color variations. Despite recent progress, existing self-supervised rPPG methods mainly reconstruct masked pixels or low-level visual representations, which can bias the model toward facial appearance rather than latent physiological dynamics. Moreover, most recent Mamba-based approaches scan facial video tokens only in chronological order, limiting their ability to exploit the cyclic structure of pulse signals. To address these limitations, we propose RhythmJEPA, a rhythm-structured joint-embedding predictive learning framework for rPPG. Instead of reconstructing RGB frames, RhythmJEPA predicts latent teacher representations from masked facial videos, thereby encouraging physiology-aware representation learning in the embedding space. To explicitly model pulse-related temporal structure, we introduce a Cyclic Rhythm-State Planner (CRSP), which estimates frame-wise latent physiological states and decodes the most plausible cyclic state path via dynamic programming with a constrained transition grammar. Guided by the decoded states, we further design a Dual-Order Mamba Encoder (DOM), which combines conventional chronological scanning with state-ordered scanning to capture both local temporal continuity and long-range rhythm-consistent dependencies. Finally, a lightweight Spatial Pulse Mixer (SPM) extracts compact pulse-sensitive facial tokens with a favorable balance between complexity and performance. Experiments on PURE, UBFC-rPPG, and MMPD show competitive performance over representative rPPG methods.
\end{abstract}
\section{Introduction}
\label{sec:intro}

\begin{figure}[t]
  \centering
  \includegraphics[width=0.8\linewidth]{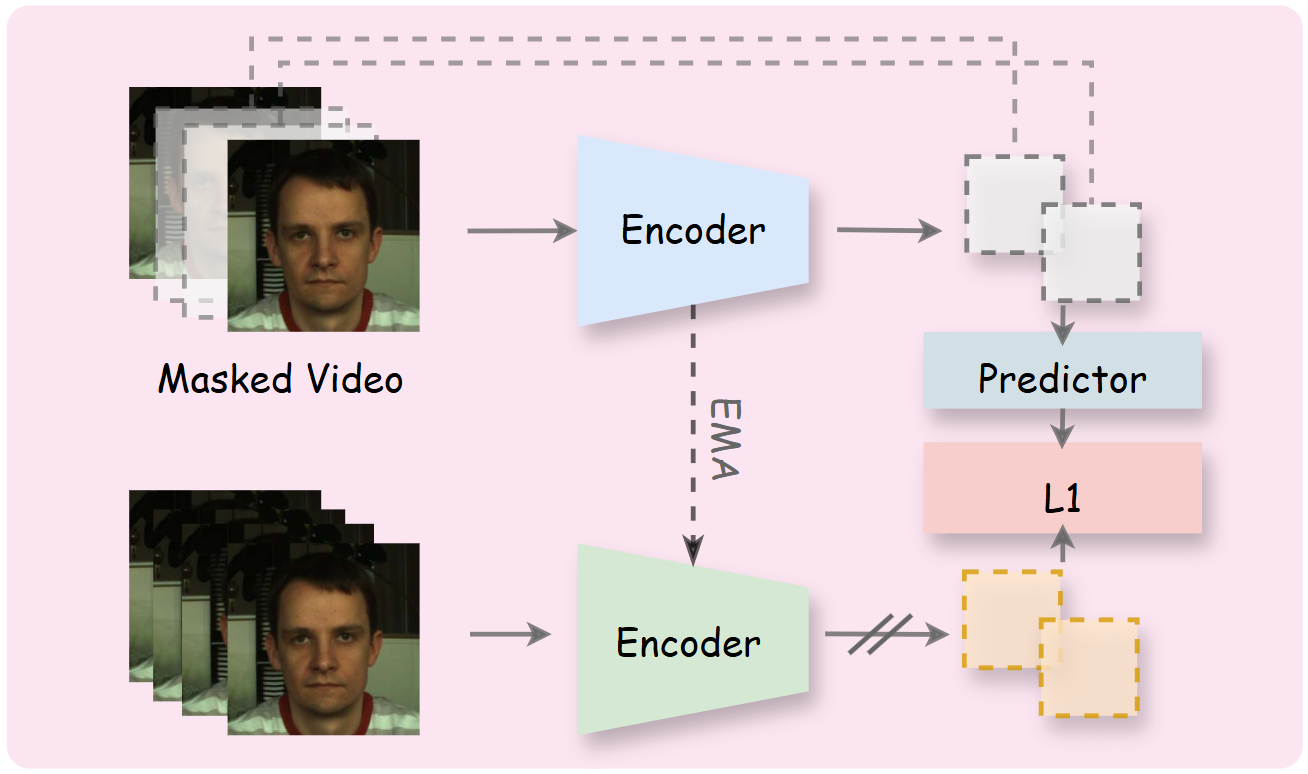}
  \caption{Stage I: Given a masked facial video, the student encoder and predictor estimate latent targets generated by an EMA teacher encoder from the original video. The model is optimized in the embedding space, avoiding direct pixel reconstruction and encouraging physiology-aware representation learning.}
  \label{fig:intro}
\end{figure}

Heart rate (HR) is a fundamental physiological indicator for health monitoring, but conventional ECG or contact PPG sensors require physical attachment and are less convenient for unobtrusive or continuous measurement. Remote photoplethysmography (rPPG) offers a camera-based alternative by estimating cardiac activity from facial videos~\cite{verkruysse2008remote}. However, reliable rPPG estimation remains challenging in unconstrained scenarios because the pulse-related visual signal is weak and easily corrupted by head motion, facial expression, illumination changes, compression artifacts, and appearance variation~\cite{huang2023challenges}.

Early rPPG methods recover pulse signals using hand-crafted signal processing or color-space priors, including ICA, CHROM, and POS~\cite{ica,chrom,pos}. These methods are interpretable and efficient, but their performance often degrades in complex real-world scenarios. Deep models further improve rPPG estimation by learning spatiotemporal representations from data~\cite{deepphys,physnet,tscan,physformer,efficientphys,rhythmformer,reperio,physmamba}. Nevertheless, most of them are trained with synchronized physiological labels and may struggle to generalize across datasets with different cameras, subjects, motions, and illumination conditions.

Self-supervised rPPG learning has recently been explored to reduce label dependence~\cite{rppg-mae,maskfusionnet,rs+,contrastphys+}. These methods learn from unlabeled facial videos through masked pre-training or strongly self-supervised objectives. However, such paradigms may still bias the learned representations toward facial appearance rather than latent physiological dynamics. Meanwhile, recent Mamba-based rPPG models further improve long-range temporal modeling with efficient state-space sequence processing~\cite{mamba,rhythmmamba,physmamba,fvbpsr_mamba}. Despite their effectiveness, they mainly process facial video tokens in chronological order, without explicitly organizing frames according to recurring physiological states. This motivates a representation learning framework that is both label-efficient and rhythm-aware. Since pulse dynamics are quasi-periodic, temporally distant frames may share similar physiological states, whereas adjacent frames can belong to different phases of the cardiac cycle. Therefore, modeling only chronological continuity may be insufficient to capture inter-period relationships that are important for robust rPPG estimation.

To address these issues, we propose \textbf{RhythmJEPA}, a rhythm-structured joint-embedding predictive learning framework for rPPG. Following I-JEPA and VideoJEPA~\cite{ijepa,videojepa}, RhythmJEPA predicts latent teacher representations from masked facial videos instead of reconstructing pixels. Since rPPG estimation can be viewed as a sequence modeling problem, we design a Spatial Pulse Mixer (SPM) as a lightweight alternative to Transformer-based spatial modeling, producing effective frame-level pulse tokens for subsequent temporal modeling. We further introduce a Cyclic Rhythm-State Planner (CRSP) to assign latent physiological states to frames, and a Dual-Order Mamba Encoder (DOM) to combine chronological scanning with rhythm-state-ordered scanning. The main contributions of this paper are summarized as follows:
\begin{itemize}
    \item We introduce RhythmJEPA, a joint-embedding predictive learning framework that learns latent physiological representations from masked facial videos for rPPG estimation.
    \item We design SPM to capture pulse-sensitive spatial cues effectively without the complexity of Transformer-based spatial modeling.
    \item We propose CRSP and DOM to model cyclic physiological recurrence beyond conventional chronological temporal scanning, together with cycle-consistency and state-balance losses that regularize rhythm-state assignments.
    \item Extensive intra- and cross-dataset experiments on PURE~\cite{pure}, UBFC-rPPG~\cite{ubfc}, and MMPD~\cite{mmpd} demonstrate the effectiveness and robustness of RhythmJEPA.
\end{itemize}

\section{Related Work}
\label{sec:relatedwork}

\subsection{Traditional rPPG Methods}

Traditional rPPG approaches recover pulse signals from facial color variations using hand-crafted physiological and optical priors. ICA and PCA separate RGB observations into latent source components~\cite{ica,pca}, while CHROM, POS, and LGI design color-space or local-invariance constraints to suppress non-pulsatile interference~\cite{chrom,pos,lgi}. Although efficient and interpretable, these methods rely on assumptions about skin reflection, illumination, and motion, which limits their robustness in unconstrained videos.

\subsection{Deep Learning-based rPPG}

Recent deep models improve rPPG estimation by learning spatiotemporal representations from facial video sequences. CNN-based and temporal-shift methods capture pulse-related appearance changes~\cite{deepphys,physnet,tscan,efficientphys}, while Transformer and periodicity-aware models enhance temporal reasoning under challenging conditions~\cite{physformer,rhythmformer,lsts,reperio}. However, most supervised approaches require synchronized physiological labels and may generalize poorly across subjects, cameras, and acquisition settings.

To reduce label dependence, self-supervised methods such as rPPG-MAE, MaskFusionNet, and RS+rPPG learn from unlabeled facial videos through masked pretraining or strong self-supervision~\cite{rppg-mae,maskfusionnet,rs+}. Nevertheless, reconstruction- or consistency-oriented objectives can still emphasize facial appearance and low-level visual variations rather than latent physiological dynamics.

\subsection{State-Space Modeling for rPPG}

State-space models provide an efficient alternative to attention-based sequence modeling. Mamba introduces input-dependent selective scanning with linear complexity, and VMamba extends this idea to visual representation learning~\cite{mamba,vmamba}. Recent rPPG models adopt Mamba-style modules for temporal physiological modeling~\cite{physmamba,rhythmmamba,fvbpsr_mamba}, but their scan orders remain largely chronological or multi-scale. RhythmJEPA instead organizes frames by latent rhythm states, enabling Mamba scanning to capture inter-period relationships across similar pulse phases.

\subsection{Joint-Embedding Predictive Learning}

Joint-embedding predictive architectures learn representations by predicting latent targets instead of reconstructing input pixels. I-JEPA demonstrates this principle for images~\cite{ijepa}, while VideoJEPA extends feature prediction to video representation learning~\cite{videojepa}. This paradigm is well suited to rPPG because the signal of interest is a latent physiological process rather than facial appearance itself. RhythmJEPA adapts joint-embedding prediction to rPPG by predicting teacher embeddings from masked facial videos and coupling this objective with rhythm-state planning and dual-order Mamba encoding.

\section{Methodology}
\label{sec:method}

\begin{figure*}[t]
    \centering
    \includegraphics[width=\linewidth]{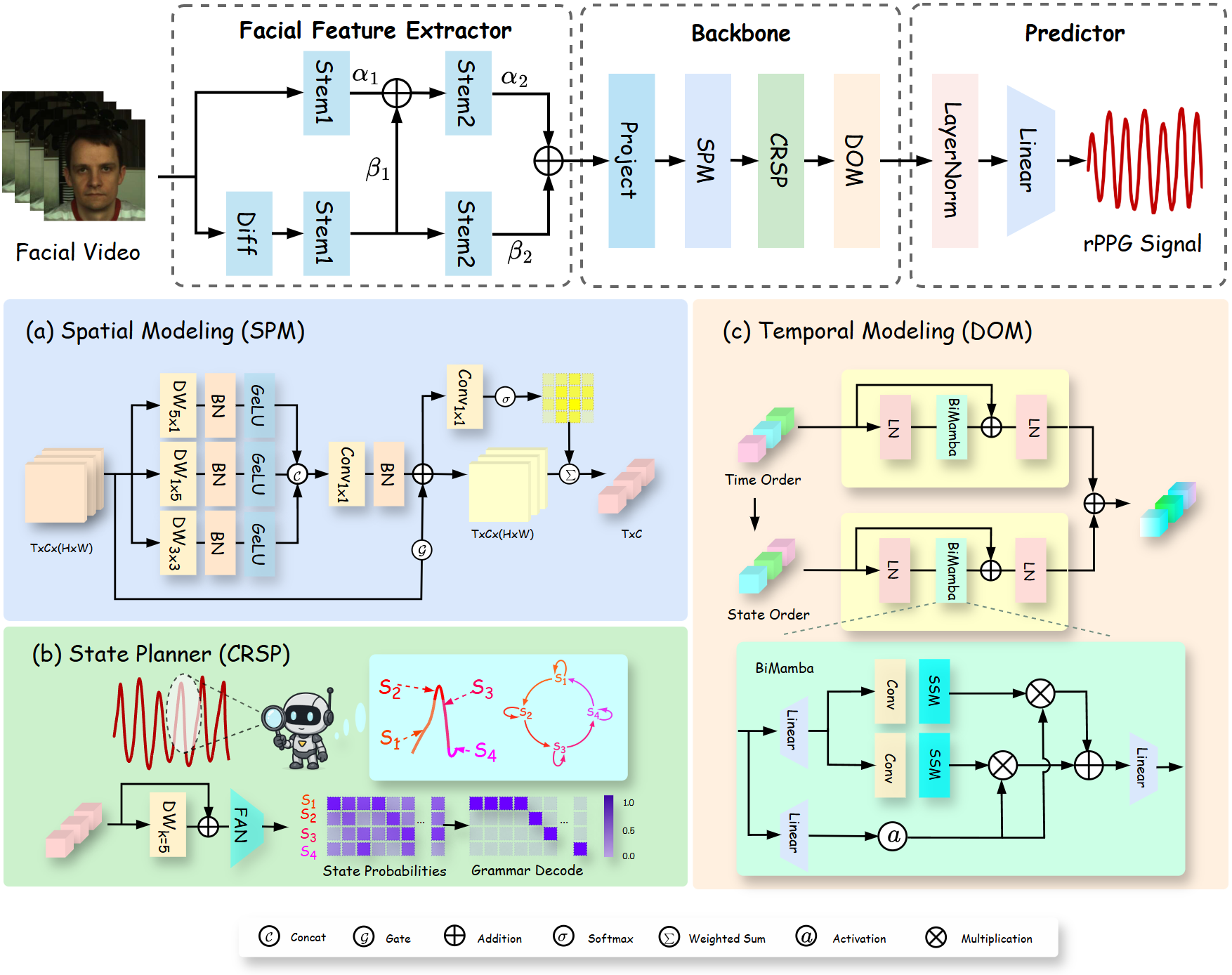}
    \caption{Stage II: RhythmJEPA fine-tuning.}
    \label{fig:rhythmjepa-finetune}
\end{figure*}

\subsection{Problem Formulation}
\label{sec:method_formulation}

We represent a facial video clip as
\begin{equation}
\mathbf{X}=\{X_t\}_{t=0}^{T-1},
\qquad
X_t\in\mathbb{R}^{H\times W\times 3}.
\end{equation}
Its synchronized rPPG waveform is represented as
\begin{equation}
\mathbf{y}=[y_0,\dots,y_{T-1}]\in\mathbb{R}^{T}.
\end{equation}
We learn a video-to-waveform mapping:
\begin{equation}
f_{\theta}:\mathbb{R}^{T\times H\times W\times 3}\rightarrow\mathbb{R}^{T},
\qquad
\hat{\mathbf{y}}=f_{\theta}(\mathbf{X}),
\end{equation}
where $\hat{\mathbf{y}}=[\hat{y}_0,\dots,\hat{y}_{T-1}]$ denotes the predicted rPPG waveform.

As shown in Fig.~\ref{fig:rhythmjepa-finetune}, RhythmJEPA consists of four modules: a facial feature encoder $E$, a Cyclic Rhythm-State Planner $\Gamma$, a Dual-Order Mamba encoder $D$, and a prediction head $h$. The overall forward process follows
\begin{equation}
\begin{aligned}
\mathbf{Z} &= E(\mathbf{X}),
&
(\mathbf{Q},\mathbf{s}^{\star}) &= \Gamma(\mathbf{Z}),\\
\mathbf{Y} &= D(\mathbf{Z},\mathbf{s}^{\star}),
&
\hat{\mathbf{y}} &= h(\mathbf{Y}).
\end{aligned}
\label{eq:rhythmjepa_forward}
\end{equation}
Here $\mathbf{Z},\mathbf{Y}\in\mathbb{R}^{T\times C}$ denote frame-wise physiological tokens and rhythm-aware temporal features, respectively. $\mathbf{Q}\in\mathbb{R}^{T\times K}$ is the rhythm-state posterior, and $\mathbf{s}^{\star}\in\{0,\ldots,K-1\}^{T}$ is the decoded cyclic state path. RhythmJEPA is first pre-trained with a rhythm-structured JEPA objective and then fine-tuned for waveform regression.

\subsection{Rhythm-Structured JEPA Pre-training}
\label{sec:method_jepa}

RhythmJEPA learns physiological representations by predicting latent teacher features instead of reconstructing RGB pixels. We sample a temporal mask $\mathbf{V}\in\{0,1\}^{T\times1\times1\times1}$, where $V_t=0$ denotes a masked frame, and define $\mathcal{M}=\{t\mid V_t=0\}$. Masked frames are replaced by a learnable RGB token $\mathbf{X}_{\mathrm{mask}}$:
\begin{equation}
\widetilde{\mathbf{X}}
=
\mathbf{V}\odot\mathbf{X}
+
(\mathbf{1}-\mathbf{V})\odot\mathbf{X}_{\mathrm{mask}}.
\end{equation}
The student branch encodes the masked clip, while the EMA teacher branch encodes the original clip:
\begin{equation}
\mathbf{R}_{s}=D_s(E_s(\widetilde{\mathbf{X}}),\mathbf{s}^{\star}_{s}),
\qquad
\mathbf{R}^{+}=\operatorname{sg}\big(D_t(E_t(\mathbf{X}),\mathbf{s}^{\star}_{t})\big).
\end{equation}
A lightweight predictor maps the student representation to the teacher latent space:
\begin{equation}
\widehat{\mathbf{R}}=P(\mathbf{R}_{s}),
\qquad
\widehat{\mathbf{R}},\mathbf{R}^{+}\in\mathbb{R}^{T\times C}.
\end{equation}
Here $\widehat{\mathbf{r}}_t$ and $\mathbf{r}^{+}_t$ denote the $t$-th temporal embeddings of $\widehat{\mathbf{R}}$ and $\mathbf{R}^{+}$, respectively. The JEPA loss is computed only at masked temporal positions:
\begin{equation}
\mathcal{L}_{\mathrm{JEPA}}
=
\frac{1}{|\mathcal{M}|}
\sum_{t\in\mathcal{M}}
\left\|
\ell_2(\widehat{\mathbf{r}}_t)
-
\ell_2(\mathbf{r}^{+}_t)
\right\|_2^2.
\end{equation}
The EMA teacher update follows
\begin{equation}
\theta_t\leftarrow \mu\theta_t+(1-\mu)\theta_s.
\end{equation}

\subsection{Facial Feature Encoder}
\label{sec:method_token_encoder}

The facial feature encoder maps each frame to a frame-level feature representation. It contains an adopted Facial Feature Extractor~\cite{qian2026rethinking} and an SPM.

\paragraph{Facial Feature Extractor.}
FFE captures local cues by combining RGB appearance with short-term temporal differences:
\begin{equation}
\mathbf{F}_{0}
=
\operatorname{FFE}(\mathbf{X}),
\qquad
\mathbf{F}_{0}\in\mathbb{R}^{T\times C_s\times H'\times W'}.
\end{equation}
A $1\times1$ convolution projects the channel dimension to $C$, followed by positional embedding:
\begin{equation}
\mathbf{F}
=
\operatorname{PE}
\left(
\phi_{1\times1}(\mathbf{F}_{0})
\right),
\qquad
\mathbf{F}\in\mathbb{R}^{T\times C\times H'\times W'}.
\end{equation}

\paragraph{Spatial Pulse Mixer.}
We propose SPM to perform lightweight spatial modeling. For a kernel size $a\times b$, the basic depth-wise block is
\begin{equation}
\mathcal{B}_{a\times b}(\mathbf{F})
=
\operatorname{GELU}
\left(
\operatorname{BN}
\left(
\operatorname{DWConv}_{a\times b}(\mathbf{F})
\right)
\right).
\end{equation}
SPM uses horizontal, vertical, and local branches:
\begin{equation}
\mathbf{F}_{h}=\mathcal{B}_{1\times5}(\mathbf{F}),
\quad
\mathbf{F}_{v}=\mathcal{B}_{5\times1}(\mathbf{F}),
\quad
\mathbf{F}_{\ell}=\mathcal{B}_{3\times3}(\mathbf{F}).
\end{equation}
Residual point-wise fusion gives the mixed feature map:
\begin{equation}
\mathbf{F}^{\star}
=
\mathbf{F}
+
\gamma\operatorname{BN}
\left(
\rho_{1\times1}
\left([
\mathbf{F}_{h};
\mathbf{F}_{v};
\mathbf{F}_{\ell}
]
\right)
\right).
\end{equation}
SPM further applies attentive spatial pooling to produce frame-level features. For each frame, the spatial attention map is computed as
\begin{equation}
\alpha_t
=
\operatorname{Softmax}_{H'W'}
\left(
\psi_{1\times1}(\mathbf{F}^{\star}_{t})
\right).
\end{equation}
The frame-level token is obtained through weighted~pooling:
\begin{equation}
\mathbf{z}_{t}
=
\sum_{i,j}
\alpha_{t,i,j}
\mathbf{F}^{\star}_{t,:,i,j}.
\end{equation}
Stacking all frame tokens yields
\begin{equation}
\mathbf{Z}
=
\operatorname{LN}
\left(
[\mathbf{z}_0,\dots,\mathbf{z}_{T-1}]
\right),
\qquad
\mathbf{Z}\in\mathbb{R}^{T\times C}.
\end{equation}

\subsection{Cyclic Rhythm-State Planner}
\label{sec:method_crsp}

CRSP estimates latent rhythm states and decodes a cyclic state path for rhythm-aware temporal scanning. Given $\mathbf{Z}$, CRSP first extracts local temporal evidence:
\begin{equation}
\mathbf{U}
=
\mathbf{Z}
+
\eta\,
\operatorname{MLP}
\left(
\operatorname{DWConv}_{5}
\left(
\mathbf{Z}
\right)
\right).
\end{equation}
A Fourier Analysis Network (FAN)~\cite{dong2026fan} predicts frame-wise rhythm-state logits:
\begin{equation}
\mathbf{R}=\operatorname{FAN}(\mathbf{U}),
\qquad
\mathbf{R}\in\mathbb{R}^{T\times K}.
\end{equation}
The rhythm-state probabilities are obtained by
\begin{equation}
\mathbf{Q}=\operatorname{Softmax}_{K}(\mathbf{R}),
\qquad
Q_{t,k}\in[0,1].
\end{equation}
Here $Q_{t,k}$ denotes the predicted probability that frame $t$ belongs to rhythm state $k$.
To better follow state transitions within the cardiac rhythm, we propose a cyclic consistency constraint that allows each state to either stay unchanged or move to the next state:
\begin{equation}
\mathcal{A}(i)=\{i,(i+1)\bmod K\}.
\end{equation}
The decoded state path is obtained from
\begin{equation}
\begin{aligned}
\mathbf{s}^{\star}
&=
\argmax_{s_{0:T-1}}
\sum_{t=0}^{T-1}
\log Q_{t,s_t},\\
\mathrm{s.t.}\quad
s_t&\in\mathcal{A}(s_{t-1}),
\quad t=1,\dots,T-1.
\end{aligned}
\end{equation}
This constrained path is obtained by Viterbi-style dynamic programming.

\paragraph{Regularization.}
Let $\mathbf{A}\in\{0,1\}^{K\times K}$ be the transition matrix induced by $\mathcal{A}$. We define the cyclic transition loss as
\begin{equation}
\mathcal{L}_{\mathrm{cyc}}
=
-
\frac{1}{T-1}
\sum_{t=0}^{T-2}
\log
\left(
\mathbf{q}_{t}^{\top}
\mathbf{A}
\mathbf{q}_{t+1}
+
\epsilon
\right),
\end{equation}
where $\mathbf{q}_{t}=\mathbf{Q}_{t,:}$. To avoid state collapse, we add the balance loss, where $\boldsymbol{\pi}$ denotes the desired prior distribution over rhythm states
\begin{equation}
\mathcal{L}_{\mathrm{bal}}
=
\operatorname{KL}
\left(
\boldsymbol{\pi}
\,\|\,
\bar{\mathbf{q}}
\right),
\end{equation}
where the average state usage is
\begin{equation}
\bar{\mathbf{q}}
=
\frac{1}{T}
\sum_{t=0}^{T-1}
\mathbf{q}_t.
\end{equation}

\subsection{Dual-Order Mamba Encoder}
\label{sec:method_dom}

Prior Mamba-based rPPG methods typically scan tokens only in chronological order, overlooking the effect of scan order on rhythm modeling. DOM addresses this by scanning the token sequence in two complementary orders. The chronological branch processes the original temporal order:
\begin{equation}
\mathbf{Y}^{\mathrm{time}}
=
\operatorname{BiMamba}_{\mathrm{time}}(\mathbf{Z}).
\end{equation}
The rhythm branch sorts tokens by decoded state and then by time:
\begin{equation}
\boldsymbol{o}
=
\operatorname{argsort}
\left(
 s^{\star}_{t}T+t
\right).
\end{equation}
The state-ordered sequence is formed as
\begin{equation}
\mathbf{Z}^{\mathrm{state}}_{r,:}
=
\mathbf{Z}_{o_r,:}.
\end{equation}
A second BiMamba scan processes this sequence:
\begin{equation}
\widetilde{\mathbf{Y}}^{\mathrm{state}}
=
\operatorname{BiMamba}_{\mathrm{state}}
\left(
\mathbf{Z}^{\mathrm{state}}
\right).
\end{equation}
We scatter the output back to the chronological order:
\begin{equation}
\mathbf{Y}^{\mathrm{state}}_{o_r,:}
=
\widetilde{\mathbf{Y}}^{\mathrm{state}}_{r,:}.
\end{equation}
A token-wise gate fuses the two temporal views:
\begin{equation}
\mathbf{G}
=
\sigma
\left(
\phi_g
\left(
[\mathbf{Z};\mathbf{Y}^{\mathrm{time}};\mathbf{Y}^{\mathrm{state}}]
\right)
\right),
\end{equation}
\begin{equation}
\mathbf{Y}
=
\mathbf{G}\odot\mathbf{Y}^{\mathrm{time}}
+
(\mathbf{1}-\mathbf{G})\odot\mathbf{Y}^{\mathrm{state}}.
\end{equation}

\subsection{Prediction Head}
\label{sec:method_objective}

The prediction head maps each fused temporal representation to a scalar pulse value:
\begin{equation}
\hat{y}_t
=
\phi_o
\left(
\operatorname{LN}(\mathbf{Y}_t)
\right).
\end{equation}
Collecting all predictions gives
\begin{equation}
\hat{\mathbf{y}}
=
[\hat{y}_0,\dots,\hat{y}_{T-1}].
\end{equation}
During fine-tuning, we optimize the negative Pearson correlation loss:
\begin{equation}
\mathcal{L}_{\mathrm{NP}}
=
1
-
\frac{
\sum_{t=0}^{T-1}
(\hat{y}_t-\bar{\hat{y}})(y_t-\bar{y})
}{
\sqrt{
\sum_{t=0}^{T-1}
(\hat{y}_t-\bar{\hat{y}})^2
}
\sqrt{
\sum_{t=0}^{T-1}
(y_t-\bar{y})^2
}
}.
\end{equation}
The overall pre-training objective combines the three losses:
\begin{equation}
\mathcal{L}_{\mathrm{pre}}
=
\lambda_{\mathrm{JEPA}}\mathcal{L}_{\mathrm{JEPA}}
+
\lambda_{\mathrm{cyc}}\mathcal{L}_{\mathrm{cyc}}
+
\lambda_{\mathrm{bal}}\mathcal{L}_{\mathrm{bal}}.
\end{equation}

\section{Experiments}
\label{sec:experiments}

This section describes the experimental protocol, including datasets, evaluation metrics, and implementation details. We then report intra-dataset and cross-dataset comparisons with representative rPPG baselines.

\subsection{Datasets}
\label{subsec:datasets}

We evaluate RhythmJEPA on three widely used rPPG benchmarks:

\textbf{UBFC-rPPG}~\cite{ubfc} contains 42 facial RGB videos recorded at 30 fps under indoor and sunlight illumination conditions. The reference physiological signals are collected by a CMS50E pulse oximeter at 60 Hz.

\textbf{PURE}~\cite{pure} includes 60 RGB videos from 10 subjects, captured at 30 fps under six predefined head-motion settings. The synchronized pulse signals are also recorded using CMS50E at 60 Hz.

\textbf{MMPD}~\cite{mmpd} is a mobile rPPG benchmark collected from 33 subjects. It includes diverse variations in skin tone, illumination, camera setting, and activity type, making it suitable for evaluating robustness under realistic domain shifts.

\definecolor{RJHead}{HTML}{DCEBFF}
\definecolor{RJSubHead}{HTML}{EDF5FF}
\definecolor{RJStripe}{HTML}{F8FBFF}
\definecolor{RJBestRow}{HTML}{FFF3C4}
\definecolor{RJRule}{HTML}{4D7DBF}

\newcommand{\best}[1]{\textbf{#1}}
\newcommand{\second}[1]{\underline{#1}}
\newcommand{\NA}{\textemdash}

\begin{table*}[t]
\centering
\scriptsize
\arrayrulecolor{RJRule}
\setlength{\tabcolsep}{2.35pt}
\renewcommand{\arraystretch}{1.08}
\resizebox{\linewidth}{!}{%
\begin{tabular}{@{}ll*{12}{c}@{}}
\toprule
\rowcolor{RJHead}
\multirow{2}{*}{\textbf{Method}} &
\multirow{2}{*}{\textbf{Venue}} &
\multicolumn{4}{c}{\textbf{PURE}} &
\multicolumn{4}{c}{\textbf{UBFC-rPPG}} &
\multicolumn{4}{c}{\textbf{MMPD}} \\
\cmidrule(lr){3-6}\cmidrule(lr){7-10}\cmidrule(l){11-14}
\rowcolor{RJSubHead}
& &
MAE$\downarrow$ & MAPE$\downarrow$ & RMSE$\downarrow$ & $r\uparrow$ &
MAE$\downarrow$ & MAPE$\downarrow$ & RMSE$\downarrow$ & $r\uparrow$ &
MAE$\downarrow$ & MAPE$\downarrow$ & RMSE$\downarrow$ & $r\uparrow$ \\
\midrule

CHROM~\cite{chrom} & TBME'13
& 5.39 & 11.08 & 15.09 & 0.81
& 4.06 & 3.34 & 8.83 & 0.89
& 13.73 & 16.95 & 18.88 & 0.15 \\

\rowcolor{RJStripe}
POS~\cite{pos} & TBME'17
& 0.36 & 0.50 & 0.93 & 1.00
& 4.08 & 3.93 & 7.72 & 0.92
& 15.61 & 18.28 & 21.40 & 0.14 \\

LGI~\cite{lgi} & CVPRW'18
& 3.59 & 3.37 & 14.66 & 0.79
& 5.39 & 11.08 & 15.09 & 0.81
& 16.63 & 18.77 & 23.06 & 0.11 \\

\rowcolor{RJStripe}
DeepPhys~\cite{deepphys} & ECCV'18
& 3.33 & 2.91 & 14.45 & 0.90
& 0.76 & 0.79 & 1.09 & 0.99
& 23.73 & 25.63 & 28.25 & -0.06 \\

PhysNet~\cite{physnet} & BMVC'19
& 0.93 & 1.40 & 2.08 & 0.99
& 2.25 & 2.37 & 4.81 & 0.94
& 4.81 & \second{4.84} & 11.83 & 0.60 \\

\rowcolor{RJStripe}
TS-CAN~\cite{tscan} & NeurIPS'20
& 0.32 & 0.50 & 0.63 & 0.99
& 1.24 & 1.35 & 2.79 & 0.96
& 8.97 & 9.43 & 16.58 & 0.44 \\

PhysFormer~\cite{physformer} & CVPR'22
& 0.52 & 0.86 & 1.03 & 0.99
& 2.34 & 2.60 & 5.55 & 0.97
& 13.64 & 14.42 & 19.39 & 0.15 \\

\rowcolor{RJStripe}
EfficientPhys~\cite{efficientphys} & WACV'23
& 0.55 & 0.71 & 1.34 & 0.99
& 0.73 & 0.83 & 2.53 & 0.97
& 12.79 & 13.48 & 21.12 & 0.24 \\

Contrast-Phys+~\cite{contrastphys+} & TPAMI'24
& 1.00 & 1.21 & 1.40 & 0.99
& 0.64 & 0.87 & 1.00 & 0.99
& 15.77 & 18.65 & 22.88 & -0.64 \\

\rowcolor{RJStripe}
RhythmMamba~\cite{rhythmmamba} & AAAI'25
& 0.29 & 0.36 & \second{0.40} & 0.99
& 0.54 & \second{0.54} & \second{0.79} & 0.99
& \second{4.12} & 4.96 & \second{9.90} & \second{0.74} \\

RhythmFormer~\cite{rhythmformer} & PR'25
& 0.27 & 0.31 & 0.46 & \second{1.00}
& 0.81 & 0.80 & 1.12 & 0.99
& 6.74 & 6.93 & 11.92 & 0.71 \\

\rowcolor{RJStripe}
LSTS~\cite{lsts} & TCSVT'25
& \second{0.15} & \second{0.22} & \second{0.40} & 0.99
& \second{0.51} & 0.55 & 1.27 & \second{0.99}
& 4.80 & 5.80 & 10.46 & 0.69 \\

\midrule
\rowcolor{RJBestRow}
\textbf{RhythmJEPA (Ours)} & \textbf{\NA}
& \best{0.12} & \best{0.18} & \best{0.28} & \best{1.00}
& \best{0.21} & \best{0.24} & \best{0.34} & \best{1.00}
& \best{3.92} & \best{4.07} & \best{9.15} & \best{0.76} \\
\bottomrule
\end{tabular}}
\caption{Intra-dataset evaluation results on PURE~\cite{pure}, UBFC-rPPG~\cite{ubfc}, and MMPD~\cite{mmpd}. Lower is better for MAE, MAPE, and RMSE, while higher is better for Pearson correlation $r$. Best results are highlighted in bold, selected second-best results are underlined.}
\label{tab:intra_dataset_comparison}
\end{table*}
\arrayrulecolor{black}

\begin{table*}[t]
\centering
\footnotesize
\arrayrulecolor{RJRule}
\setlength{\tabcolsep}{3.5pt}
\renewcommand{\arraystretch}{1.15}
\resizebox{\linewidth}{!}{%
\begin{tabular}{l*{16}{c}}
\toprule
\rowcolor{RJHead}
\multirow{2}{*}{\textbf{Model}} &
\multicolumn{4}{c}{PURE $\rightarrow$ UBFC-rPPG} &
\multicolumn{4}{c}{UBFC-rPPG $\rightarrow$ PURE} &
\multicolumn{4}{c}{PURE $\rightarrow$ MMPD} &
\multicolumn{4}{c}{UBFC-rPPG $\rightarrow$ MMPD} \\
\cmidrule(lr){2-5}\cmidrule(lr){6-9}\cmidrule(lr){10-13}\cmidrule(lr){14-17}
\rowcolor{RJSubHead}
& MAE$\downarrow$ & MAPE$\downarrow$ & RMSE$\downarrow$ & $r\uparrow$
& MAE$\downarrow$ & MAPE$\downarrow$ & RMSE$\downarrow$ & $r\uparrow$
& MAE$\downarrow$ & MAPE$\downarrow$ & RMSE$\downarrow$ & $r\uparrow$
& MAE$\downarrow$ & MAPE$\downarrow$ & RMSE$\downarrow$ & $r\uparrow$ \\
\midrule

DeepPhys~\cite{deepphys}
& 1.21 & 1.42 & 2.90 & 0.99
& 8.06 & 13.67 & 19.71 & 0.61
& 16.92 & 18.54 & 24.61 & 0.05
& 17.50 & 19.27 & 25.00 & 0.05 \\

\rowcolor{RJStripe}
PhysNet~\cite{physnet}
& \underline{0.98} & \underline{1.12} & \underline{2.48} & \underline{0.99}
& \underline{3.69} & \underline{3.39} & \underline{13.80} & \underline{0.82}
& \underline{13.22} & \underline{14.73} & \underline{19.61} & \underline{0.23}
& \underline{10.24} & \underline{12.46} & \underline{16.54} & \underline{0.29} \\

TSCAN~\cite{tscan}
& 1.30 & 1.50 & 2.87 & 0.99
& 12.92 & 23.92 & 24.36 & 0.47
& 13.94 & 15.14 & 21.61 & 0.20
& 14.01 & 15.48 & 21.04 & 0.24 \\

\rowcolor{RJStripe}
PhysFormer~\cite{physformer}
& 1.44 & 1.66 & 3.77 & 0.98
& 12.92 & 23.92 & 24.36 & 0.47
& 14.57 & 16.73 & 20.71 & 0.15
& 12.10 & 15.41 & 17.79 & 0.17 \\

EfficientPhys~\cite{efficientphys}
& 2.07 & 2.10 & 6.32 & 0.94
& 5.47 & 5.40 & 17.04 & 0.71
& 14.03 & 15.32 & 21.62 & 0.17
& 13.78 & 15.15 & 22.25 & 0.09 \\

\rowcolor{RJStripe}
RhythmFormer~\cite{rhythmformer}
& 2.26 & 2.34 & 5.48 & 0.95
& 8.02 & 15.71 & 18.14 & 0.74
& 17.52 & 20.20 & 25.51 & 0.18
& 14.95 & 20.46 & 21.16 & 0.18 \\

\rowcolor{RJBestRow}
\textbf{RhythmJEPA (Ours)}
& \textbf{0.71} & \textbf{1.04} & \textbf{1.94} & \textbf{1.00}
& \textbf{3.07} & \textbf{3.38} & \textbf{8.93} & \textbf{0.92}
& \textbf{9.46} & \textbf{10.79} & \textbf{16.40} & \textbf{0.27}
& \textbf{9.15} & \textbf{11.79} & \textbf{14.80} & \textbf{0.31} \\
\bottomrule
\end{tabular}}
\caption{Cross-dataset test results. PURE$\rightarrow$UBFC-rPPG means the model is trained on PURE and tested on UBFC-rPPG. Lower is better for MAE/MAPE/RMSE, while higher is better for Pearson correlation $r$.}
\label{tab:cross_comparison}
\end{table*}
\arrayrulecolor{black}

\subsection{Evaluation Metrics}
\label{subsec:eval_metrics}

Following mainstream evaluation settings~\cite{efficientphys,rhythmformer,reperio}, we evaluate HR estimation using mean absolute error (MAE), mean absolute percentage error (MAPE), root mean square error (RMSE), and Pearson's correlation coefficient ($r$). MAE/RMSE measure absolute errors, while MAPE reflects relative error. Pearson's $r$ highlights the linear association between predicted and ground-truth HRs, where higher values indicate better consistency.

\subsection{Implementation Details}
\label{subsec:impl_details}

Face regions are first detected by RetinaFace~\cite{retinaface}. Each bounding box is enlarged by a factor of \(1.5\), cropped, and resized to \(128\times128\) pixels. Training clips contain \(180\) frames with a stride of \(90\), and random horizontal flipping is used for augmentation.

The token dimension is set to \(C=96\), and RhythmJEPA uses \(4\) latent rhythm states. The state planner employs FAN with hidden, periodic, and non-periodic dimensions of \(96\), \(24\), and \(48\), respectively. The context encoder uses dual-order Bi-Mamba scanning with \(3\) blocks per scan and Mamba state dimension \(16\). In the JEPA stage, the teacher momentum and mask ratio are set to \(0.996\) and \(0.7\), respectively. The loss weights are set to \(\lambda_{\mathrm{JEPA}}=1.0\), \(\lambda_{\mathrm{cyc}}=0.1\), and \(\lambda_{\mathrm{bal}}=0.1\).

All models are trained on an NVIDIA GeForce RTX 3090 GPU using PyTorch 2.0.0 with CUDA 11.8. We use AdamW with batch size \(4\) and an initial learning rate of \(1\times10^{-4}\). The learning rate follows a OneCycle schedule~\cite{Onecycle} with cosine annealing. Pretraining is conducted for \(30\) epochs, while finetuning is performed for \(10\) epochs.

\subsection{Main Comparison}
For intra-dataset evaluation, we follow the training and testing splits used in LSTS~\cite{lsts} to ensure a fair comparison with prior rPPG methods. For cross-dataset evaluation, the model is trained on the source dataset and directly tested on the target dataset without using target-domain training samples.

\subsubsection{Intra-dataset Testing}
Tab.~\ref{tab:intra_dataset_comparison} summarizes the within-dataset evaluation on PURE~\cite{pure}, UBFC-rPPG~\cite{ubfc}, and MMPD~\cite{mmpd}. RhythmJEPA ranks first across all reported metrics, while the underlined entries indicate the strongest competing results. On PURE, it improves over LSTS by 20.0\%/18.2\%/30.0\% in MAE/MAPE/RMSE and reaches a perfect correlation of 1.00, matching the best baseline correlation. On UBFC-rPPG, the error reduction is more substantial: RhythmJEPA lowers the second-best MAE/MAPE/RMSE from 0.51/0.54/0.79 to 0.21/0.24/0.34, corresponding to 58.8\%/55.6\%/57.0\% relative improvements. On MMPD, which contains stronger appearance, illumination, and activity variations, RhythmJEPA also reduces MAE/MAPE/RMSE by 4.9\%/15.9\%/7.6\% over the best competing error results and improves the correlation from 0.74 to 0.76. These consistent gains indicate that rhythm-structured predictive learning helps recover pulse-related cues more reliably across both clean and challenging intra-dataset settings.

\subsubsection{Cross-dataset Testing}
Tab.~\ref{tab:cross_comparison} further examines cross-dataset transfer, where training and testing videos come from different acquisition domains. RhythmJEPA achieves the best results in all four transfer directions, showing consistent relative error reductions over the strongest baseline. For PURE$\rightarrow$UBFC-rPPG, RhythmJEPA reduces MAE/MAPE/RMSE by 27.6\%/7.1\%/21.8\% over PhysNet, while also increasing the correlation from 0.99 to 1.00. For UBFC-rPPG$\rightarrow$PURE, the improvement is especially clear in RMSE, decreasing from 13.80 to 8.93, with the highest correlation of 0.92. When transferring to the more challenging MMPD dataset, RhythmJEPA reduces MAE/MAPE/RMSE by 28.4\%/26.7\%/16.4\% for PURE$\rightarrow$MMPD and by 10.6\%/5.4\%/10.5\% for UBFC-rPPG$\rightarrow$MMPD. The correlation also improves from 0.23 to 0.27 and from 0.29 to 0.31, respectively. These results suggest that RhythmJEPA learns physiological representations that transfer more effectively across datasets, rather than relying primarily on dataset-specific appearance statistics.
\subsubsection{Computational Cost and Inference Efficiency}
Table~\ref{tab:efficiency} compares the computational cost and inference speed of RhythmJEPA against representative rPPG baselines under the same benchmarking setting. RhythmJEPA uses only 3.20M parameters and 33.50G MACs, making it substantially lighter than recent Transformer-based methods such as PhysFormer~\cite{physformer}, LSTS~\cite{lsts}, and Reperio-rPPG~\cite{reperio}. Although EfficientPhys~\cite{efficientphys} is the smallest and fastest model, its accuracy is clearly lower than RhythmJEPA in Tables~\ref{tab:intra_dataset_comparison} and~\ref{tab:cross_comparison}, especially on challenging MMPD settings. In contrast, RhythmJEPA achieves the best estimation accuracy while retaining the second-best parameter count, MACs, and latency in Table~\ref{tab:efficiency}, demonstrating a favorable accuracy--efficiency trade-off.

\section{Ablation Study}
\label{sec:ablation}

This section analyzes the core design choices of RhythmJEPA from spatial, temporal, and pre-training perspectives. We first examine whether SPM attends to physiologically meaningful facial regions, then compare different Mamba scan orders to isolate the effect of rhythm-state scanning. Finally, we investigate why JEPA pre-training is useful for learning rhythm-aware representations before supervised rPPG fine-tuning.

\subsection{Analysis of Spatial Pulse Modeling}
\label{subsec:ablation_spm}

\begin{figure*}[t]
\centering
\includegraphics[width=0.98\textwidth]{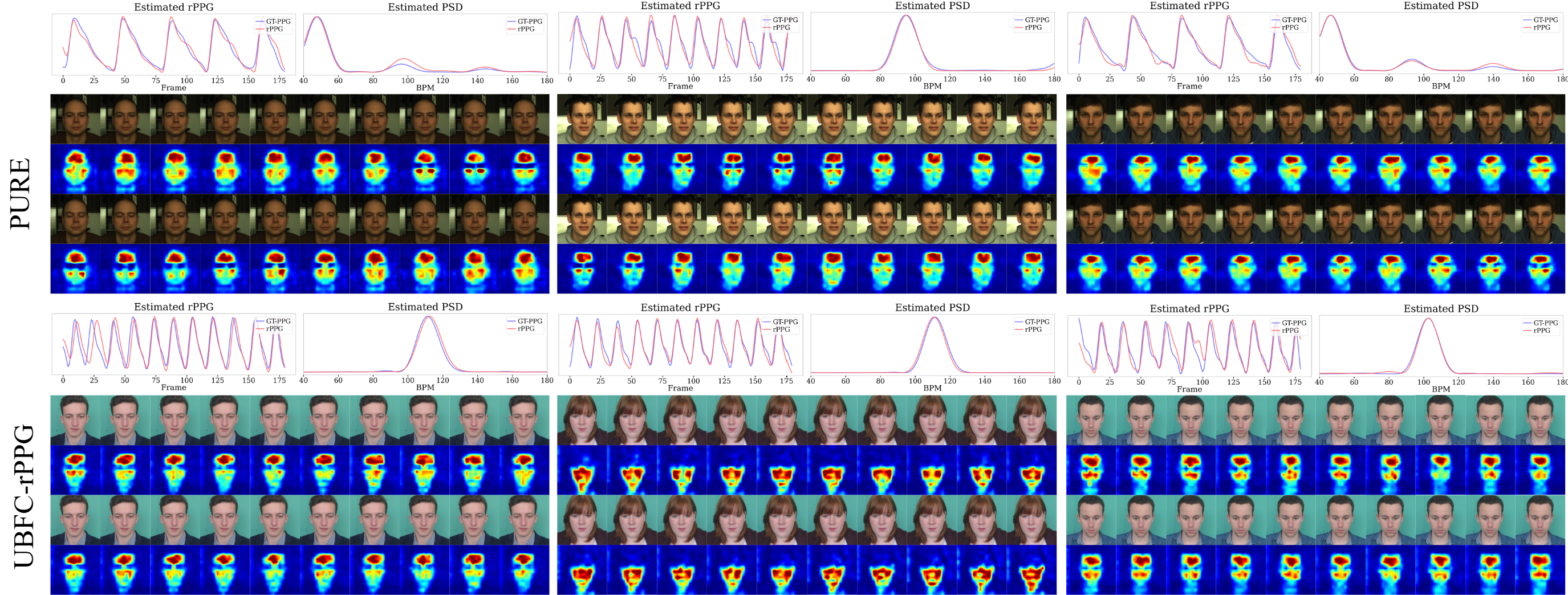}
\caption{Qualitative analysis of SPM on PURE and UBFC-rPPG. We jointly visualize the recovered rPPG waveform, its power spectrum, and spatial heatmaps to examine whether SPM emphasizes pulse-informative facial regions while producing physiologically consistent predictions.}
\label{fig:spm_analysis}
\end{figure*}

Figure~\ref{fig:spm_analysis} analyzes the behavior of SPM on PURE and
UBFC-rPPG using three complementary views: the recovered rPPG waveform, the
corresponding power spectrum, and spatial heatmaps. The predicted waveform is
highly correlated with the ground-truth BVP signal, closely matching its periodic
peaks and valleys. The power spectrum further shows that the dominant frequency
of the prediction aligns well with the reference rhythm, indicating accurate and
temporally consistent pulse recovery.

The heatmaps further show that SPM suppresses background or weakly informative
regions and instead highlights facial areas such as the cheeks, forehead, and
nose. Since these regions typically exhibit stronger pulse-induced color
variations, this pattern suggests that SPM provides a pulse-aware spatial prior
rather than relying on irrelevant appearance or background cues.

\subsection{Effect of Rhythm-State Mamba Scanning}
\label{subsec:ablation_scan}

We first examine whether the gain comes from the Mamba temporal module alone
or from the proposed rhythm-structured scan order. The chronological scan
processes frames in their original temporal order, whereas the rhythm-state scan
reorders frames according to the latent rhythm states decoded by CRSP. As shown
in Table~\ref{tab:scan_ablation}, replacing chronological scanning with
rhythm-state scanning reduces the error on both PURE and MMPD, indicating that
frames sharing similar pulse states provide useful long-range dependencies even
when they are not adjacent in time. The dual-order scan further combines local
temporal continuity with rhythm-consistent dependencies and achieves the best
MAE and correlation on both datasets, as well as the best RMSE on MMPD. On PURE,
the rhythm-state-only scan obtains the lowest RMSE, suggesting that this cleaner
setting benefits strongly from phase-consistent grouping, while dual-order
scanning remains more balanced across metrics and datasets.

\begin{table}[t]
\centering
\caption{Ablation of Mamba scan strategies.}
\label{tab:scan_ablation}
\resizebox{\linewidth}{!}{%
\begin{tabular}{lcccccc}
\toprule
\multirow{2}{*}{Temporal modeling} & \multicolumn{3}{c}{PURE} & \multicolumn{3}{c}{MMPD} \\
\cmidrule(lr){2-4} \cmidrule(lr){5-7}
 & MAE $\downarrow$ & RMSE $\downarrow$ & $r \uparrow$ & MAE $\downarrow$ & RMSE $\downarrow$ & $r \uparrow$ \\
\midrule
Chronological scan & 0.26 & 0.42 & 0.99 & 5.10 & 10.63 & 0.71 \\
Rhythm-state scan & 0.15 & \textbf{0.26} & 0.99 & 4.86 & 9.80 & 0.73 \\
Dual-order scan & \textbf{0.12} & 0.28 & \textbf{1.00} & \textbf{3.92} & \textbf{9.15} & \textbf{0.76} \\
\bottomrule
\end{tabular}}

\end{table}

\begin{figure}[t]
\centering
\includegraphics[width=\linewidth]{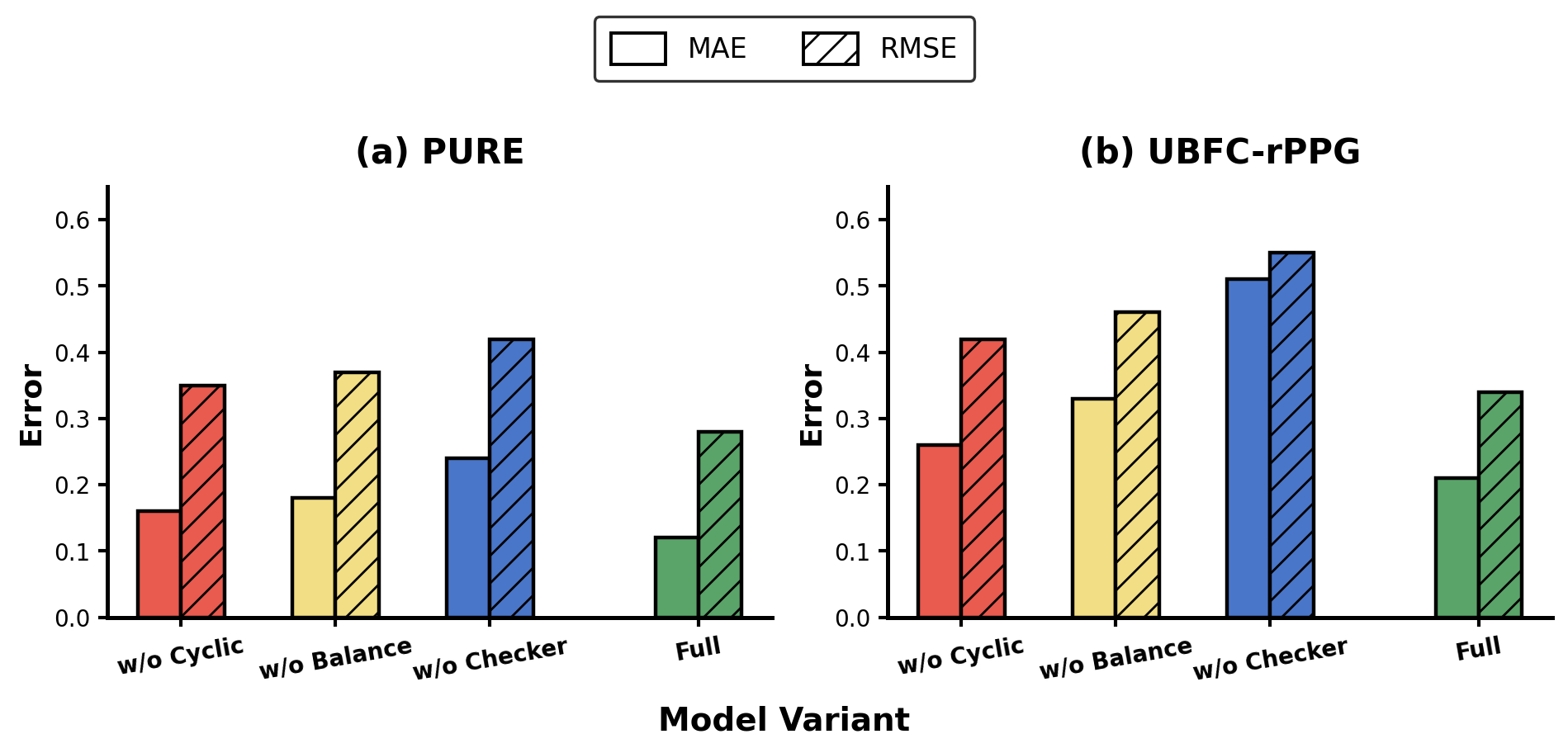}
\caption{Ablation of CRSP constraints for rPPG estimation. The bar plots compare the full model with variants that remove the cyclic-transition loss, remove the state-balance loss, or remove the grammar checker. Assigning states only from frame-wise probabilities or weakening the rhythm-state regularization produces less structured state sequences and degrades performance.}
\label{fig:grammar_checker_ablation}
\end{figure}

Table~\ref{tab:scan_ablation} also depends on whether CRSP can produce
physiologically meaningful state sequences for the rhythm-state scan. We
therefore ablate the CRSP constraints in Figure~\ref{fig:grammar_checker_ablation}.
Removing the grammar checker assigns each frame directly from frame-wise softmax
probabilities, which can make the decoded state path noisy, disordered, and
inconsistent with the cyclic pulse progression assumed by the scan. Removing the
cyclic-transition loss weakens the phase-order prior, while removing the
state-balance loss makes the learned states less evenly used across the pulse
cycle. In all cases, performance drops compared with the full model. Together,
Table~\ref{tab:scan_ablation} and Figure~\ref{fig:grammar_checker_ablation}
show that RhythmJEPA benefits not only from scanning in rhythm-state order, but
also from learning stable, balanced, and cyclically valid state assignments before
performing that scan.

\subsection{Analysis of JEPA Pretraining}
\label{subsec:ablation_jepa}

We next isolate the contribution of joint-embedding predictive pretraining.
Both variants use the same downstream architecture and supervised fine-tuning
protocol; they differ only in whether the model is initialized by RhythmJEPA
pretraining. As shown in Table~\ref{tab:pretraining_ablation}, pretraining
reduces MAE and RMSE on all three datasets. The gain is modest on PURE, where the
baseline is already strong, but becomes clearer on UBFC-rPPG and MMPD. In
particular, pretraining improves UBFC-rPPG from 0.36/0.51 to 0.21/0.34 in
MAE/RMSE and improves MMPD from 5.02/10.36 to 3.92/9.15. These results suggest
that predicting masked teacher embeddings provides a useful rhythm-aware
initialization, especially when appearance, illumination, and motion variations
make supervised fine-tuning more difficult.

\begin{table}[t]
\centering
\caption{Ablation of JEPA pretraining. The check mark indicates whether the model is initialized with the proposed RhythmJEPA pretraining before supervised rPPG fine-tuning.}
\label{tab:pretraining_ablation}
\resizebox{\linewidth}{!}{%
\begin{tabular}{cccccccccc}
\toprule
\multirow{2}{*}{JEPA} & \multicolumn{3}{c}{PURE} & \multicolumn{3}{c}{UBFC-rPPG} & \multicolumn{3}{c}{MMPD} \\
\cmidrule(lr){2-4} \cmidrule(lr){5-7} \cmidrule(lr){8-10}
 & MAE $\downarrow$ & RMSE $\downarrow$ & $r \uparrow$ & MAE $\downarrow$ & RMSE $\downarrow$ & $r \uparrow$ & MAE $\downarrow$ & RMSE $\downarrow$ & $r \uparrow$ \\
\midrule
$\times$ & 0.15 & 0.31 & 1.00 & 0.36 & 0.51 & 0.99 & 5.02 & 10.36 & 0.72 \\
$\checkmark$ & \textbf{0.12} & \textbf{0.28} & \textbf{1.00} & \textbf{0.21} & \textbf{0.34} & \textbf{1.00} & \textbf{3.92} & \textbf{9.15} & \textbf{0.76} \\
\bottomrule
\end{tabular}}
\end{table}

\begin{figure}[t]
\centering
\includegraphics[width=\linewidth]{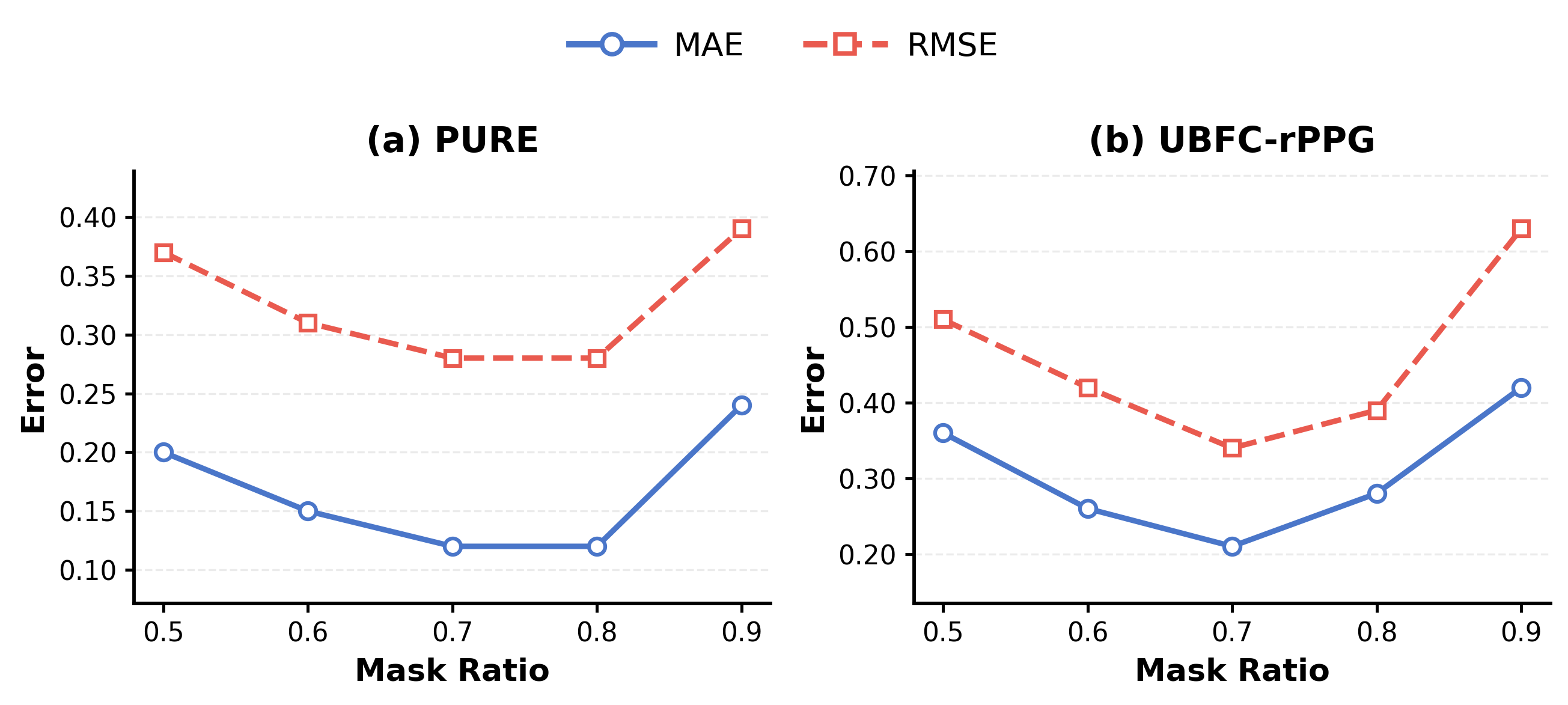}
\caption{Sensitivity analysis of the JEPA mask ratio on PURE and UBFC-rPPG. All variants use the same fine-tuning protocol; only the mask ratio during pretraining is changed.}
\label{fig:mask_ratio_ablation}
\end{figure}

\begin{table}[t]
\centering
\caption{Efficiency comparison against representative baselines. We report the number of parameters, computational cost measured by MACs, throughput, and inference latency under the same benchmarking setting.}
\label{tab:efficiency}
\setlength{\tabcolsep}{3.2pt}
\renewcommand{\arraystretch}{1.08}
\resizebox{\linewidth}{!}{%
\begin{tabular}{lcccc}
\toprule
Method & \#Param (M)$\downarrow$ & MACs (G)$\downarrow$ & Throughput (Kfps)$\uparrow$ & Latency (ms)$\downarrow$ \\
\midrule
MTTS-CAN~\cite{tscan} & 4.33 & 45.85 & \underline{3.99} & 50.03 \\
EfficientPhys~\cite{efficientphys} & \textbf{2.16} & \textbf{22.96} & \textbf{7.21} & \textbf{27.72} \\
PhysFormer~\cite{physformer} & 7.38 & 60.72 & 2.27 & 84.50 \\
LSTS~\cite{lsts} & 6.37 & 71.06 & 1.30 & 147.68 \\
Reperio-rPPG~\cite{reperio} & 6.50 & 71.02 & 1.47 & 130.18 \\
\textbf{RhythmJEPA (Ours)} & \underline{3.20} & \underline{33.50} & 3.66 & \underline{49.15} \\
\bottomrule
\end{tabular}}
\end{table}
We further analyze the mask ratio used during JEPA pretraining in
Figure~\ref{fig:mask_ratio_ablation}. The experiment is conducted on PURE and
UBFC-rPPG while keeping the fine-tuning protocol fixed, so the curves reflect the
sensitivity of the learned initialization to the pretext-task difficulty. On
PURE, increasing the mask ratio from 0.5 to 0.7--0.8 steadily reduces both MAE
and RMSE, but the error rises again at 0.9. UBFC-rPPG shows a similar trend, with
the best performance at 0.7 and a clear degradation when the ratio is too high.
This pattern indicates that a low mask ratio gives an overly easy prediction
task, whereas an excessive ratio removes too much temporal evidence for stable
rhythm prediction. A moderate-to-high ratio therefore provides the best balance
between pretraining difficulty and physiological recoverability.

\section{Conclusion}
\label{sec:Conclusion}
In this paper, we presented RhythmJEPA, a joint-embedding predictive learning framework for rPPG that focuses on learning physiological representations rather than reconstructing facial appearance. By predicting teacher embeddings from masked facial videos, the model is encouraged to capture information related to the underlying pulse signal while reducing dependence on pixel-level reconstruction. We also introduced SPM to extract frame-level representations with a lightweight spatial design, avoiding the computational burden of Transformer-based spatial modeling. To better exploit the periodic nature of rPPG signals, CRSP estimates latent rhythm states for video frames, and DOM uses these states to scan the sequence in both chronological and rhythm-aware orders. This design allows the model to consider not only neighboring frames but also frames that may correspond to similar phases of the pulse cycle. Experiments on PURE, UBFC-rPPG, and MMPD show that RhythmJEPA improves both intra-dataset accuracy and cross-dataset generalization over representative baselines. Despite these results, the learned rhythm states are still implicit and may not directly correspond to clinically interpretable cardiac phases. Future work will investigate more interpretable rhythm modeling and evaluate the framework under more diverse real-world conditions.
{
    \small
    \bibliographystyle{ieeenat_fullname}
    \bibliography{main}
}

\end{document}